\documentclass[runningheads]{llncs}

\usepackage{eccv}

\usepackage{eccvabbrv}

\usepackage{graphicx}
\usepackage{booktabs}
\usepackage{multirow}
\usepackage[accsupp]{axessibility}  
\usepackage{comment}

\usepackage{hyperref}

\usepackage{orcidlink}

\def\methodname{OP-Align}

\begin{document}

\title{\methodname: Object-level and Part-level Alignment for Self-supervised Category-level Articulated Object Pose Estimation} 

\titlerunning{\methodname: Object-level and Part-level Alignment}

\author{Yuchen Che\inst{1}\orcidlink{0009-0007-9055-4726} \and
Ryo Furukawa\inst{2}\orcidlink{0009-0008-6862-0800} \and
Asako Kanezaki\inst{1}\orcidlink{0000-0003-3217-1405}}

\authorrunning{Y.~Che et al.}

\institute{
    Tokyo Institute of Technology, Tokyo, Japan\\ 
    \email{cheyuchen.titech@gmail.com, kanezaki@c.titech.ac.jp} \and
    Accenture Japan Ltd, Tokyo, Japan\\
    \email{rfurukaward@gmail.com}
}

\maketitle

\begin{abstract}
Category-level articulated object pose estimation focuses on the pose estimation of unknown articulated objects within known categories. Despite its significance, this task remains challenging due to the varying shapes and poses of objects, expensive dataset annotation costs, and complex real-world environments.
In this paper, we propose a novel self-supervised approach that leverages a single-frame point cloud to solve this task.
Our model consistently generates reconstruction with a canonical pose and joint state for the entire input object, and it estimates object-level poses that reduce overall pose variance and part-level poses that align each part of the input with its corresponding part of the reconstruction.
Experimental results demonstrate that our approach significantly outperforms previous self-supervised methods and is comparable to the state-of-the-art supervised methods.
To assess the performance of our model in real-world scenarios, we also introduce a new real-world articulated object benchmark dataset\footnote{Code and dataset are released at \href{https://github.com/YC-Che/OP-Align}{https://github.com/YC-Che/OP-Align}.}.
\end{abstract}

\section{Introduction}
\begin{figure}[t]
\begin{minipage}[b]{0.58\textwidth}
    \centering
    \scalebox{0.82}{
    \begin{tabular}{@{}ccccc@{}}
        \toprule
        \multirow{2}{*}{Method} & w/o Pose & w/o Shape & Single & Real-time\\
         & Supervision & Supervision & Frame  & Inference\\
        \cmidrule(lr){1-1} \cmidrule(lr){2-5}
        PartMobility~\cite{partMobility}  & \checkmark & \checkmark & & \\
        UPPD~\cite{kawana}  & \checkmark & & \checkmark  & \checkmark\\
        EAP~\cite{PLSEE} & \checkmark & \checkmark  & \checkmark \\
        Ours & \checkmark & \checkmark & \checkmark  & \checkmark\\
        \bottomrule
    \end{tabular}
    }
    \captionof{table}{Overview of works on self-supervised category-level articulated object pose estimation.} 
    \label{tab:previous_work_overview}
\end{minipage}
\hspace{3mm}%
\begin{minipage}[b]{0.39\textwidth}
    \centering%
    \includegraphics[width=.95\linewidth]{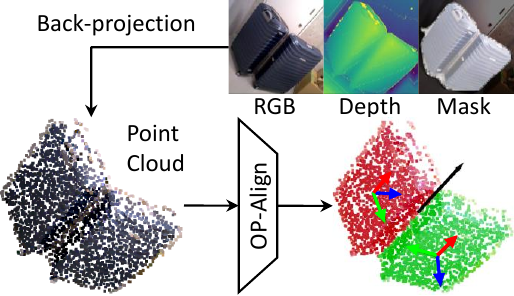}
    \caption{Illustration of the articulated object pose estimation.}
    \label{fig:objective}
\end{minipage}
\end{figure}

Articulated objects, comprising multiple parts connected by revolute or prismatic joints with varying joint states (rotational angle of a revolute joint or translation length of a prismatic joint), commonly exist in the real world. The interactions between humans and these objects give rise to numerous practical applications, such as robot manipulations and automation in industrial processes~\cite{command,hoi4d}. Therefore, pose estimation for these objects has become a crucial problem in computer vision.
We focus on accomplishing the category-level articulated object pose estimation through a self-supervised approach. Our objective is to use a point cloud of unknown articulated objects within known categories obtained from a single-frame RGB-D image segmented by detection models such as Mask-RCNN~\cite{MaskRCNN} as input. Then, we infer each part's pose and segmentation, each joint's direction and pivot, as illustrated in Fig.~\ref{fig:objective}. We aim to achieve this \emph{without utilizing pose and shape annotations} during training. Due to the varying shapes, poses, and complex real-world environments, this task is ill-posed and remains challenging.

Many works have focused on solving the aforementioned task under simpler problem settings. Unsupervised Pose-aware Part Decomposition~(UPPD)~\cite{kawana} utilizes object shape annotations as a substitute for pose annotations. PartMobility~\cite{partMobility} utilizes multiple-frame point clouds of the same object under different joint states. However, these methods still face limitations when confronted with scenarios where shape information is unavailable or when dealing with single-frame data. To the best of our knowledge, Equi-Articulated-Pose~(EAP)~\cite{PLSEE} is the only work that has tackled this task with single-frame point cloud as input and without shape or pose annotations on a synthetic dataset.
EAP guides the network to learn part-by-part reconstruction of the input shapes by combining disentangled information, such as canonical part shapes, object structure, and part-level poses, in a self-supervised manner.
To achieve such disentanglement, EAP extracts part-level $\mathrm{SE}(3)$-equivariant shape feature of a local region, instead of object-level $\mathrm{SE}(3)$-equivariant one, from an input and part-level poses.
Since part-level poses are not given in nature, EAP requires iterative updates of such poses.
It also uses an inner iterative operation, Slot-Attention~\cite{slot-attention}, for segmenting parts.
These iterative operations sacrifice inference speed.

We propose Object-level and Part-level Alignment~(\methodname), a novel self-supervised approach that learns object-level alignment, part-level alignment, and canonical reconstruction of the entire object rather than the part-by-part reconstructions.
The core idea is that part segmentation and part-level pose estimation should be done for objects with low object-level pose variance.
Based on this idea, we reconsider the order of the process of the part-by-part reconstruction approach (EAP) and propose a new learning strategy.
In our approach, the network first generates a reconstruction that maintains the canonical pose and joint state for the entire input and aligns the input with the reconstruction at the object-level to reduce the overall pose variance.
Then, the network segments parts followed by aligning each part of the object-level aligned input and the corresponding part of the reconstruction by simulating joint movement.
Our approach does not employ iterative operation, thus achieving real-time inference speed.
A comparison with previous works is presented in Tab.~\ref{tab:previous_work_overview}.

We compare \methodname\ with other methods on a synthetic dataset. To further test \methodname's performance, we generate a real-world RGB-D dataset with multiple categories of articulated objects. 
Experimental results demonstrate that our approach achieves state-of-the-art performance with other self-supervised methods and comparable performance with other \emph{supervised} methods on the synthetic dataset and the real-world dataset while achieving real-time inference speed. 

Our contributions are summarized as follows:
\begin{enumerate}
    \item We propose a new model designed for category-level articulated object pose estimation in a self-supervised manner, which requires no of pose or shape annotations.
    \item We generate a new real-world RGB-D dataset for the category-level articulated object pose estimation.
    \item We conduct experiments on a synthetic dataset and our real-world dataset. Our model achieves comparable performance with the state-of-the-art supervised methods and significantly outperforms previous self-supervised methods while achieving real-time inference speed.
\end{enumerate}

\section{Related Works}
\textbf{Category-level rigid object pose estimation:} This task focuses on predicting an unknown rigid object's pose from images. NOCS~\cite{NOCS} predicts the per-pixel coordinates in canonical space from RGB-D images.
Several methods~\cite{Shapo, CenterSnap, ShapePrior} further employ CAD models from ShapeNet dataset~\cite{ShapeNet} to generate shape templates and use iterative closest point~(ICP)~\cite{ICP} for matching the pose.
Commonly used backbone for this task is 3D graph convolution network~(3DGCN)~\cite{3DGCN} and PointNet++~\cite{pointnet++}. These methods require expensive large-scale dataset annotation.
Some approaches attempt to accomplish this task in a self-supervised manner. With the CAD model available, several methods~\cite{AAE, self6d} render the predicted pose with the CAD model as a synthetic image and compare it with the input image. Some methods focus on the multi-view RGB images provided cases~\cite{multiplePCPose, Multi-view}. Especially, $\mathrm{SE}(3)$-eSCOPE~\cite{SE3Rigid} achieved this task with single-view input and without pose annotations or CAD models. They use the $\mathrm{SE}(3)$-equivariant backbone, Equivariant Point Net~(EPN)~\cite{EPN}, to simultaneously conduct $\mathrm{SE}(3)$-invariant shape reconstruction as a reference frame, and predict the $\mathrm{SE}(3)$-equivariant pose transformation which sends input to the reconstruction.

\textbf{Category-level articulated object pose estimation:} This task focuses on predicting part-level pose, part-level segmentation, and joint information for unknown objects within known categories. Previous methods~\cite{ANCSH, opd, generalize_kinematic, captra} try to solve this task with RGB-D image or video input by directly estimating the part-level pose. Some methods~\cite{asdf, cadex, neuralPart} transfer the task into a movable shape reconstruction task with neural implicit representation~\cite{OccNet, DeepSDF} and predict the pose indirectly. Some methods~\cite{active_articulation, ditto} parameterize the joint movement with active interaction with articulated objects. To reduce the segmentation cost, \cite{semi_augmentation} uses semantic segmentation annotation and transfers it into part segmentation to conduct semi-supervised learning. However, similar to the rigid object pose estimation, the cost of the dataset annotation limits the application of these methods.
To solve this task in a self-supervised manner, UPPD~\cite{kawana} utilizes the annotation of object shape instead of the annotation of object pose. Some methods~\cite{partMobility, multibody} used multi-view observation with the same object in different joint states to predict the joint movement. EAP~\cite{PLSEE} solved such a task with a single-frame point cloud input and without shape or pose annotation. EAP repeats the process of segmenting each part, reconstructing the per-part $\mathrm{SE}(3)$-invariant shape, and predicting the per-part pose multiple times to gain a refined pose estimation.
However, directly segmenting parts for inputs with different poses and shapes is challenging, often resulting in poor accuracy, and the inference speed is unsuitable for real-time applications.

\textbf{Articulated Object Dataset:} Synthetic datasets of articulated objects such as Shape2Motion, SAPIEN, and PartNet~\cite{shape2motion, SAPIEN, partnet} are commonly used in the articulated object pose estimation.
Compared to RGB-D images captured from the real world, these datasets lack the consideration of complicated real-world environments.
HOI4D~\cite{hoi4d} collects multiple articulated and rigid object mesh data and RGB-D images in human-object interaction. However, due to the mismatch between the depth and RGB channels, a non-negligible amount of noise is present in their ground-truth annotation of part segmentation based solely on the RGB channels.

\section{Method}


\begin{figure}[t]
    \centering
    \includegraphics[width=\linewidth]{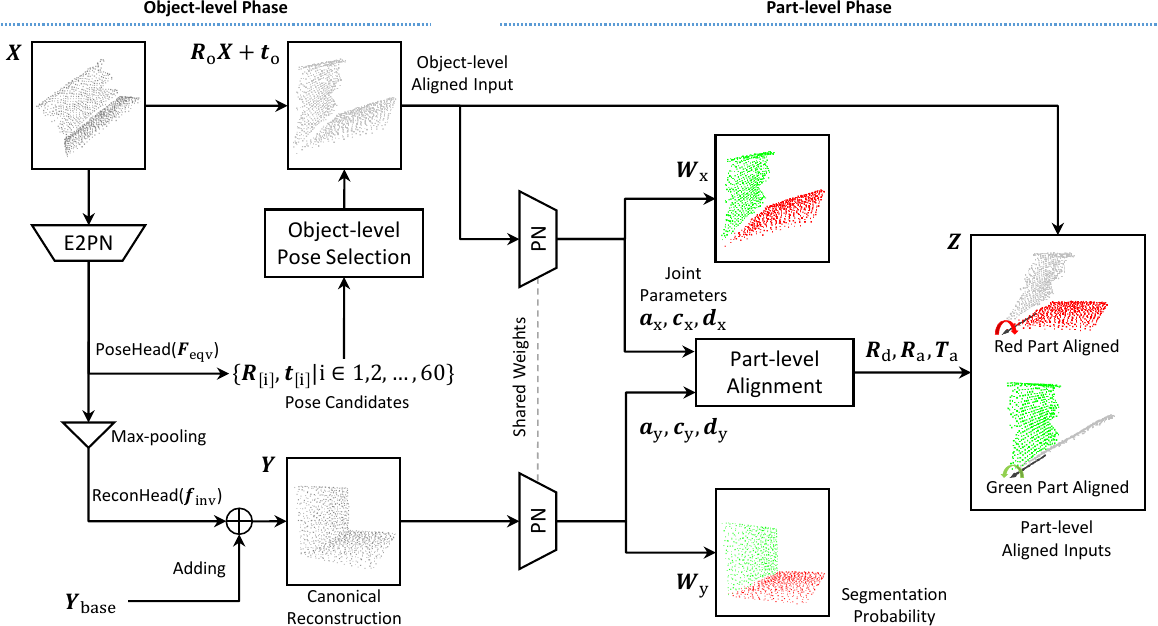}
    \caption{Pipeline of \methodname. At the object-level phase, for the input point cloud $\mathbf{X}$, we use the E2PN~\cite{e2pn} backbone to predict and select object-level pose $\mathbf{R}_{\mathrm{o}}, \mathbf{t}_{\mathrm{o}}$ from pose candidates, and generate the canonical reconstruction $\mathbf{Y}$ by adding a learnable parameter called category-common base shape $\mathbf{Y}_{\mathrm{base}}$. At the part-level phase, two PointNets~\cite{PointNet} with shared weights predict the part segmentation probability $\mathbf{W}_{\mathrm{x}}, \mathbf{W}_{\mathrm{y}}$, joint states $\mathbf{a}_{\mathrm{x}}, \mathbf{a}_{\mathrm{y}}$, joint pivots $\mathbf{c}_{\mathrm{x}}, \mathbf{c}_{\mathrm{y}}$, and joint directions $\mathbf{d}_{\mathrm{x}}, \mathbf{d}_{\mathrm{y}}$ for object-level aligned input $\mathbf{R}_{\mathrm{o}}\mathbf{X} + \mathbf{t}_{\mathrm{o}}$ and reconstruction $\mathbf{Y}$, to generate part-level alignment $\mathbf{R}_{\mathrm{d}}, \mathbf{R}_{\mathrm{a}}, \mathbf{T}_{\mathrm{a}}$ that aligns each part of $\mathbf{X}$ to the corresponding part of $\mathbf{Y}$ as part-level aligned inputs $\mathbf{Z}$.}
    \label{fig:pipeline}
\end{figure}

Category-level articulated object pose estimation can be defined as follows.
Given a point cloud $\mathbf{X} \in \mathbb{R}^{3 \times N}$ of an articulated object consisting of $P$ parts, we assign each point to a part, predict the rotation and translation for each part, and provide the pivot and the direction for each joint.
To solve this problem, our model predicts each point's segmentation probability $\mathbf{W} \in  \mathbb{R}^{P \times N}$, each joint's pivot and direction $\mathbf\{ \mathbf{c}_{[i]} \in \mathbb{R}^{3}, \mathbf{d}_{[i]} \in \mathbb{R}^{3} \mid i\in\{1,2,\dots,J\}\}$, and the rotation and the translation for each part $\{ \mathbf{R}_{[i]} \in \mathrm{SO}(3), \mathbf{t}_{[i]} \in \mathbb{R}^{3} \mid i\in\{1,2,\dots,P\}\}$.
During training, we assume that the number of parts $P$ and the type of joints (revolute or prismatic) are given. Specifically, \methodname\ assumes that each joint connects two independent parts, resulting in $J = P-1$ joints, which cover most of the articulated object categories found in daily environments.

The pipeline of \methodname\ is shown in Fig.~\ref{fig:pipeline}.
At the object-level phase, \methodname\ initially employs Efficient $\mathrm{SE}(3)$-equivariant Point Net~(E2PN)~\cite{e2pn} for object-level pose selection from a discretization of the $\mathrm{SE}(3)$ group, and generate canonical reconstruction.
At the part-level phase, two PointNets (PNs)~\cite{PointNet} with shared weights perform part segmentation and joint parameters estimation separately for the input aligned with object-level pose and the canonical reconstruction.
The obtained joint parameters generate the part-level alignment between the input and the canonical reconstruction, aligning each part of the input with its corresponding part of the reconstruction by simulating the joint movement.

In Section~\ref{preliminaries}, we will introduce the concept of object-level and part-level alignment and the required weighted point cloud distance for training. Then we will introduce the object-level phase and part-level phase of our model in Section~\ref{object-level_recon_and_pose} and Section~\ref{part-level_joint_transform}.

Notice in this section, for a rank $n$ tensor $A$, we denote the $(i_1,i_2,\dots,i_n)$-element (a rank $0$-tensor) as $A_{[i_1,i_2,\dots,i_n]}$. 
Moreover, we use NumPy~\cite{numpy} like notation to extract a tensor from $A$ (but each index starts from $1$).
For example, $A_{[i_1]}$ denote the $i_1$-th rank $(n-1)$ tensor along the first axis and $A_{[:, i_2]}$ denote the $i_2$-th rank $(n-1)$ tensor along the second axis.

\subsection{Preliminaries}\label{preliminaries}

\begin{figure}[t]
    \centering
    \includegraphics[width=0.95\linewidth]{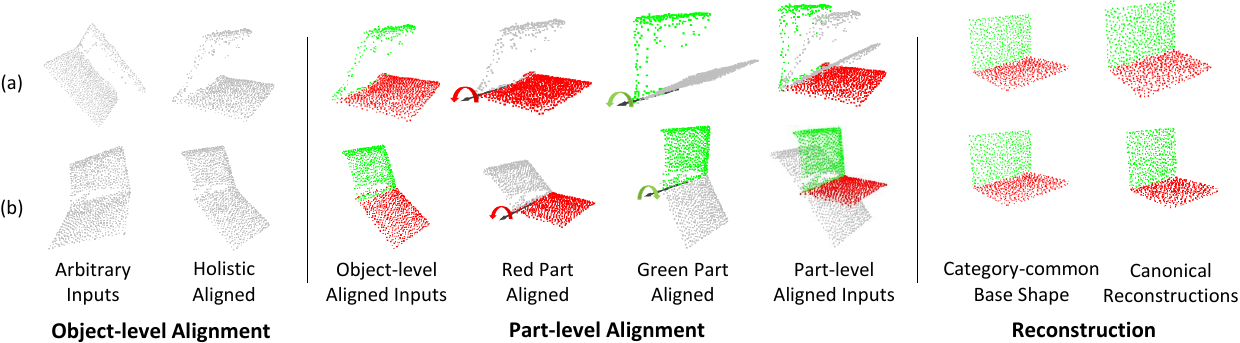}
    \caption{Illustration of the object-level alignment, part-level alignment, and the reconstruction of two inputs (a) and (b). Object-level alignment aligns the inputs with the canonical reconstructions holistically. Part-level alignment simulates joint movement to align each part. The category-common base shape remains consistent for all inputs, and the canonical reconstruction further fits the shape details of each input.}
    \label{fig:variance}
\end{figure}

\subsubsection{Expansion from rigid objects to articulated objects}
To solve the rigid object pose estimation in a self-supervised manner, $\mathrm{SE}(3)$-eSCOPE~\cite{SE3Rigid} utilizes a $\mathrm{SE}(3)$-equivariant backbone to disentangle shape and pose by generating an $\mathrm{SE}(3)$-invariant shape reconstruction and selecting $\mathrm{SE}(3)$-equivariant pose from candidates in a discretization of the $\mathrm{SE}(3)$ group for aligning the reconstruction and input.
They observed that poses of $\mathrm{SE}(3)$-invariant reconstructions for objects in the same category are often consistently aligned.
However, for articulated objects, each part's pose is also influenced by joint movement.
This complexity renders the reconstruction generated by the $\mathrm{SE}(3)$-eSCOPE unable to maintain consistent poses for all the parts.

To extend such an approach to articulated objects, as depicted in Fig.~\ref{fig:variance}, we use object-level alignment to reduce the overall pose variance, part-level alignment to simulate joint movement and align each part, and generate reconstruction with canonical pose and joint state for any objects.
Specifically, For object-level alignment, we use a similar strategy with $\mathrm{SE}(3)$-eSCOPE, by selecting the pose generating the smallest point cloud distance between the reconstruction and input, among multiple pose candidates, in other words, anchors. For part-level alignment, we collectively align each part of the input to the corresponding part of the reconstruction by aligning joint direction and pivot, then rotate/translate the input along the joint direction, to obtain multiple part-level aligned inputs each of which is aligned only with the corresponding part of the reconstruction. It is essential to note that each 
 part-level aligned input also leaves other parts unaligned. We use this phenomenon and calculate each point's distance between each part-level aligned input and the reconstruction to determine whether a point in each part-level aligned input belongs to the currently aligned part which guides the part segmentation learning.
To stabilize the reconstruction, we add a category-common base shape as learnable parameters to represent a common shape of all the objects in the same category.

\subsubsection{Weighted Point Cloud Distances}
We combine part segmentation probability with the point cloud distance between the part-level aligned inputs and the reconstruction to learn part segmentation and part alignment simultaneously.
To achieve this, we use weighted point cloud distances, and later, part segmentation probability will sometimes be set as weights.
A commonly used point cloud distance is the chamfer distance~(CD), and we also employ the Density-awarded Chamfer Distance~(DCD)~\cite{DCD}. 
Given two point clouds $\mathbf{P}$ and $\mathbf{Q}$, the single-directional weighted CD (L1) and DCD from $\mathbf{P}$ to $\mathbf{Q}$ with the weight $\mathbf{w}$ are defined as
\begin{equation}
\begin{split}
    \mathrm{CD}(\mathbf{P}, \mathbf{Q},  \mathbf{w}) = & \frac{1}{|\mathbf{P}|} \sum_{n=1}^{|\mathbf{P}|} \mathbf{w}_{[n]} \min_{m \in \{ 1, 2, \dots, |\mathbf{Q}|\}} \left\|\mathbf{P}_{[n]} - \mathbf{Q}_{[m]}\right\|, \\
    \mathrm{DCD}(\mathbf{P}, \mathbf{Q}, \mathbf{w}, \alpha) = & \frac{1}{|\mathbf{P}|} \sum_{n=1}^{|\mathbf{P}|}  \mathbf{w}_{[n]} \min_{m \in \{ 1, 2, \dots, |\mathbf{Q}|\}} \left( 1 - e^{-\alpha \left\|\mathbf{P}_{[n]} - \mathbf{Q}_{[m]}\right\|_2}\right). \\
\end{split}
\end{equation}
The sensitive distance range of DCD can be adjusted with the hyper-parameter $\alpha$.

\subsection{Object-level phase}\label{object-level_recon_and_pose}
In the Object-level phase, \methodname\ performs object-level pose selection, following a methodology similar to $\mathrm{SE}(3)$-eSCOPE~\cite{SE3Rigid}, and generate canonical reconstruction.

By feeding the input $\mathbf{X}$, E2PN~\cite{e2pn} backbone initially outputs the $\mathrm{SE}(3)$-equivariant feature $\mathbf{F}_{\mathrm{eqv}} \in \mathbb{R}^{D \times 60}$. This feature is generated by $60$ anchors representing different poses of the object.
Here, $60$ is the number of elements of the icosahedral rotation group, a discretization of the $3$D rotation group $\mathrm{SO}(3)$.
Then, we max pool $\mathbf{F}_{\mathrm{eqv}}$ among anchor dimension to obtain a $\mathrm{SE}(3)$-invariant feature $\mathbf{f}_{\mathrm{inv}} \in \mathbb{R}^{D}$.
We use \textrm{PoseHead}, consisting of multi-layer perceptron~(MLP), to output per-anchor rotation and translation $\{(\mathbf{R}_{[i]}, \mathbf{t}_{[i]}) = \textrm{PoseHead}(\mathbf{F}_{\mathrm{eqv}[:,i]})\mid i\in \{1,2,\dots,60\}\}$.
To obtain the canonical reconstruction $\mathbf{Y} \in \mathbb{R}^{3 \times N}$, we also use an MLP called $\textrm{ReconHead}$ and a learnable parameter $\mathbf{Y}_\mathrm{base}$ which represents the category-common base shape and is of the same size as $\mathbf{Y}$.
The canonical reconstruction $\mathbf{Y}$ is obtained by adding the output of $\textrm{ReconHead}$ and $\mathbf{Y}_\mathrm{base}$; $\mathbf{Y}=\textrm{ReconHead}(\mathbf{f}_{\mathrm{inv}}) + \mathbf{Y}_\mathrm{base}$.

We also need to select the correct object-level pose from per-anchor rotation and translation $\{(\mathbf{R}_{[i]}, \mathbf{t}_{[i]})\}$.
We calculate the single-directional CD between the input transformed by the rotation and the translation of each anchor and the reconstruction.
Then we select the anchor's rotation and translation that minimize CD as an object-level pose;
\begin{equation}
\mathbf{R}_{\mathrm{o}},\ \mathbf{t}_{\mathrm{o}} = \underset{i \in \{1,2,\dots,60\}}{\operatorname{argmin}} \mathrm{CD}(\mathbf{R}_{[i]}\mathbf{X} + \mathbf{t}_{[i]}, \mathbf{Y}, \mathbf{1}),
\end{equation}
where $\mathbf{1}$ represents the vector with all elements equal to $1$.
Notice that we do not expect the object-level pose $\mathbf{R}_{\mathrm{o}}$, $\mathbf{t}_{\mathrm{o}}$ obtained here to be accurate because we have not considered joint movement in this phase of the model.
However, $\mathbf{R}_{\mathrm{o}}$, $\mathbf{t}_{\mathrm{o}}$ can reduce the overall pose variance for subsequent non-$\mathrm{SE}(3)$-equivariant model's inputs by applying $\mathbf{R}_{\mathrm{o}}\mathbf{X} + \mathbf{t}_{\mathrm{o}}$ as object-level aligned input.

\subsubsection{Object-level Losses}
We employ DCD as the object-level reconstruction loss
\begin{equation}
    \mathcal{L}_{\mathrm{o}} = \mathrm{DCD}(\mathbf{R}_{\mathrm{o}}\mathbf{X} + \mathbf{t}_{\mathrm{o}}, \mathbf{Y}, \mathbf{1}, \alpha_{\mathrm{L}}) + \mathrm{DCD}(\mathbf{Y}, \mathbf{R}_{\mathrm{o}}\mathbf{X} + \mathbf{t}_{\mathrm{o}}, \mathbf{1}, \alpha_{\mathrm{R}}),
\end{equation}
where $\alpha_{\mathrm{L}} = 30$ and $\alpha_{\mathrm{R}} = 120$.

In addition, two regularization losses are introduced to make reconstructions more stable.
The first one is for shape variance between category-common base shape $\mathbf{Y}_{\mathrm{base}}$ and canonical reconstruction $\mathbf{Y}$ and is defined by
\begin{equation}
    \mathcal{L}_{\mathrm{regS}} = \frac{1}{N} \sum_{i=1}^{N} \left\|\mathbf{Y}_{[i]} - \mathbf{Y}_{\mathrm{base}[i]}\right\|_2.
\end{equation}
The second one is a local density regularization to ensure the reconstruction does not contain outliers and avoids sparse density in certain parts. It is defined by
\begin{equation}
    \mathcal{L}_{\mathrm{regD}} = \frac{1}{K-1} \sum_{i = 2}^{K} \mathrm{Var}(\|\mathbf{Y} - \mathrm{KNN}(\mathbf{Y}, k)\|),
\end{equation}
where $\mathrm{KNN}(\mathbf{Y}, k)$ refers to the $k$-th nearest point from each point in $\mathbf{Y}$, and we set $K = 64$ in this paper.

\subsection{Part-level phase}\label{part-level_joint_transform}

In this phase, we focus on segmenting both the object-level aligned input and the reconstruction into parts and estimating their joint parameters.
By comparing the obtained joint pivots, joint directions, and joint states, we determine the relative pose transformations to align each part of the input with the corresponding part of the reconstruction.

\begin{figure}[t]
    \centering
    \includegraphics[width=0.76\linewidth]{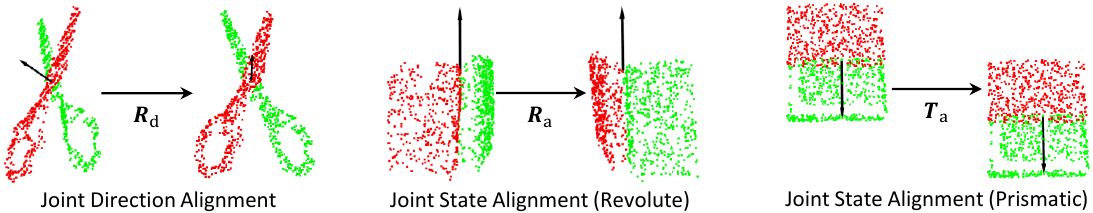}
    \caption{Illustration of joint direction alignment $\mathbf{R}_{\mathrm{d}}$, joint state alignment $\mathbf{R}_{\mathrm{a}}$ that simulating revolute joint movement, and $\mathbf{t}_{\mathrm{a}}$ that simulating prismatic joint movement.}
    \label{fig:joint_alignment}
\end{figure}

\methodname\ uses two PNs~\cite{PointNet} with shared weights to process the object-level aligned input $\mathbf{R}_{\mathrm{o}}\mathbf{X} + \mathbf{t}_{\mathrm{o}}$ and the reconstruction $\mathbf{Y}$ separately.
These two PNs output the segmentation probabilities $\mathbf{W}_\mathrm{x}, \mathbf{W}_\mathrm{y} \in \mathbb{R}^{P \times N}$, joint pivots $\mathbf{c}_\mathrm{x}, \mathbf{c}_\mathrm{y} \in \mathbb{R}^{(P-1) \times 3}$, joint directions $\mathbf{d}_\mathrm{x}, \mathbf{d}_\mathrm{y} \in \mathbb{R}^{(P-1) \times 3}$ and per-part joint states $\mathbf{a}_\mathrm{x}, \mathbf{a}_\mathrm{y} \in \mathbb{R}^{(P-1) \times 2}$ from each joint, where subscripts $\mathrm{x}$ and $\mathrm{y}$ indicate outputs from $\mathbf{R}_{\mathrm{o}}\mathbf{X} + \mathbf{t}_{\mathrm{o}}$ and $\mathbf{Y}$ respectively.
Here, joint state $\mathbf{a}_*$ represents joint angles for revolute joints and translation lengths for prismatic joints, and the dimension of the second axis of $\mathbb{R}^{(P-1) \times 2}$ reflects the assumption that each joint connect two parts.
We define the part-level aligned inputs $\mathbf{Z}_{[j,i]}$, $j=1,2,\dots,P-1$, $i=1,2$, obtained by a relative transformation that aligns the $i$-th part connected to the $j$-th joint of $\mathbf{R}_{\mathrm{o}}\mathbf{X} + \mathbf{t}_{\mathrm{o}}$ with the corresponding part of $\mathbf{Y}$ by
\begin{equation}
\label{eqn:Z}
\mathbf{Z}_{[j,i]} = 
\begin{cases}
        \mathbf{R}_{\mathrm{a}[j,i]} \mathbf{R}_{\mathrm{d}[j]} ((\mathbf{R}_{\mathrm{o}} \mathbf{X} + \mathbf{t}_{\mathrm{o}}) - \mathbf{c}_{\mathrm{x}[j]}) + \mathbf{c}_{\mathrm{y}[j]}& \text{(revolute joint)},\\
        \mathbf{R}_{\mathrm{d}[j]} ((\mathbf{R}_{\mathrm{o}} \mathbf{X} + \mathbf{t}_{\mathrm{o}}) - \mathbf{c}_{\mathrm{x}[j]}) + \mathbf{c}_{\mathrm{y}[j]} + \mathbf{T}_{\mathrm{a}[j,i]} & \text{(prismatic joint)}.
\end{cases}
\end{equation}
Here, $\mathbf{R}_{\mathrm{d}[j]}$ is a rotation matrix of the joint direction alignment that sends the joint direction $\mathbf{d}_{\mathrm{x}[j]}$ to $\mathbf{d}_{\mathrm{y}[j]}$; $\mathbf{R}_{\mathrm{d}[j]} \mathbf{d}_{\mathrm{x}[j]} = \mathbf{d}_{\mathrm{y}[j]}$.
$\mathbf{R}_{\mathrm{a}[j,i]}$ is the rotation matrix of joint state alignment, the rotation of a revolute joint with rotation angle $\mathbf{a}_{\mathrm{y}[j,i]}- \mathbf{a}_{\mathrm{x}[j,i]}$ around the axis $\mathbf{d}_{\mathrm{y}[j]}$.
And $\mathbf{T}_{\mathrm{a}[j,i]}$ is the joint state alignment translation $\mathbf{d}_{\mathrm{y}[j]} (\mathbf{a}_{\mathrm{y}[j,i]} - \mathbf{a}_{\mathrm{x}[j,i]})$ which represents a translation of a prismatic joint. The illustration of such alignments are shown in Fig.~\ref{fig:joint_alignment}.
By applying the above equation to each part, \methodname\ generates a point cloud set, part-level aligned inputs $\mathbf{Z}=\{ \mathbf{Z}_{[j,i]} \mid i\in \{1,2\}, j\in \{1,2,\dots,P-1\}\}$ where each part of the input $\mathbf{X}$ is aligned to the corresponding part of the reconstruction $\mathbf{Y}$.

\subsubsection{Corresponding part assignment}
Objects with more than two parts, such as \texttt{eyeglasses} or \texttt{basket}, have some shared parts, each of which is connected with multiple joints. These shared parts result in the number of part-level aligned inputs $\mid \mathbf{Z} \mid$ not necessarily being the same as the number of parts $P$.
To correlate $\mathbf{Z}$ with part segmentation probability, we assign one part label $\sigma(j,i)$ to each pair of a joint $j$ and a part $i$ connected to this joint, $j=1,2,\dots,P-1$, $i=1,2$.
We require the assignment $\sigma$ to satisfy two conditions: $(1)$ for any $j \in \{1,2,\dots,P-1\}$ $\sigma(j,1) \ne \sigma(j,2)$ and $(2)$ for any $p \in \{1,2,\dots,P\}$ there exist $j$ and $i$ such that $\sigma(j,i)=p$.
Let $\sigma$ be the assignment that minimizes the (sum of) segmentation-weighted CD calculated by $\sum_{j} \sum_{i} \frac{1}{\mathbf{b}_{[j,i]}} \mathrm{CD}(\mathbf{Z}_{[j,i]}, \mathbf{Y}, \mathbf{W}_{\mathrm{x}[\sigma(j,i)]})$ among all possible assignments satisfying $(1)$ and $(2)$. Here $\mathbf{b}_{[j,i]}$ denotes the number of times the part $\sigma(j,i)$ is shared.
During the inference phase, we use the mean translation by linear interpretation and the mean rotation by the spherical linear interpolation (SLERP) as the shared part's pose.

\subsubsection{Part-level Losses}
We employ a segmentation-weighted DCD as the part-level reconstruction loss
\begin{equation}
\begin{split}
    \mathcal{L}_{\mathrm{p}} = \sum_{j=1}^{P-1} \sum_{i=1}^{2} \frac{1}{\mathbf{b}_{[j,i]}} (\mathrm{DCD}(\mathbf{Z}_{[j,i]}, \mathbf{Y}, \mathbf{W}_{\mathrm{x}[\sigma(j,i)]},  \alpha_{\mathrm{L}}) + \mathrm{DCD}(\mathbf{Y}, \mathbf{Z}_{[j,i]}, \mathbf{W}_{\mathrm{y}[\sigma(j,i)]}, \alpha_{\mathrm{R}})).
\end{split}
\end{equation}

We also add some regularization. We assume that the mean segmentation probability of each part exceeds the threshold $\beta$ in the reconstruction and apply the segmentation regularization by
\begin{equation}
    \mathcal{L}_{\mathrm{regW}} = \frac{1}{P} \sum_{p=1}^{P}  \max \left(\beta -  \frac{\sum_{i=1}^{N} \mathbf{W}_{\mathrm{y}[p,i]}}{N}, 0\right),
\end{equation}
where $\beta$ is set to $0.05$.
we consider that the part-level aligned inputs of one shared part should coincide and introduce a regularization loss $\mathcal{L}_{\mathrm{regP}}$;
\begin{equation}
    \mathcal{L}_{\mathrm{regP}} = \frac{1}{2(P-1)} \sum_{j=1}^{P-1} \sum_{i=1}^{2} \left\|\mathbf{Z}_{[j,i]} - \overline{\mathbf{Z}_{[j,i]}}\right\|_2,
\end{equation}
where $\overline{\mathbf{Z}_{[j,i]}}$ indicates the mean shape of $\{\mathbf{Z}_{[a,b]} | \sigma(a,b) = \sigma(j,i)\}$.
And since the reconstruction should have a fixed canonical joint state, we define the reconstruction $\mathbf{Y}$'s joint state $\mathbf{a}_{\mathrm{y}}$ as zero and apply the joint state regularization by
\begin{equation}
    \mathcal{L}_{\mathrm{regA}} = \frac{1}{2(P-1)} \sum_{j=1}^{P-1} \sum_{i=1}^{2} \mathbf{a}_{\mathrm{y}[j,i]}^2.
\end{equation}
Finally, since both the predicted joint pivots of the input and that of the reconstruction should be close to the object itself, we applied a regularization defined by
\begin{equation}
    \mathcal{L}_{\mathrm{regJ}} = \mathrm{DCD}(\mathbf{c}_{\mathrm{y}[j]}, \mathbf{Y}, \mathbf{1}, \alpha_{\mathrm{L}}) + \mathrm{DCD}(\mathbf{c}_{\mathrm{x}[j]}, \mathbf{R}_{\mathrm{o}}\mathbf{X} + \mathbf{t}_{\mathrm{o}}, \mathbf{1}, \alpha_{\mathrm{R}}),
\end{equation}
as the joint pivot regularization.

\section{Real-world Dataset}

\begin{table}[tp]
    \centering
    \caption{Overview of the real-world dataset. The real-world dataset contains object categories with different number of parts, number of joints, and joint types.} 
    \scalebox{0.8}{
    \begin{tabular}{@{}ccccccccc@{}}
        \toprule
        \multirow{2}{*}{Category}   &\multicolumn{2}{c}{Training} &\multicolumn{2}{c}{Testing} & \multicolumn{3}{c}{Object} & \multirow{2}{*}{Detection}\\
                                    &Image &Instance              &Image &Instance             & Part & Joint(prismatic) & Joint(revolute) & \\
        \cmidrule(lr){1-1} \cmidrule(lr){2-3} \cmidrule(lr){4-5} \cmidrule(lr){6-8} \cmidrule(lr){9-9}
        \texttt{basket} & 974 & 4 & 449 & 2 & 3 & 0 & 2 & SAM~\cite{sam}\\
        \texttt{drawer} & 884 & 4 & 452 & 2 & 2 & 1 & 0 & SAM~\cite{sam}\\
        \texttt{laptop}  & 740 & 4 & 412 & 2 & 2 & 0 & 1 & Mask-RCNN~\cite{MaskRCNN}\\
        \texttt{scissors}  & 922 & 4 & 421 & 2 & 2 & 0 & 1 & Mask-RCNN~\cite{MaskRCNN}\\
        \texttt{suitcase} & 813 & 4 & 381 & 2 & 2 & 0 & 1 & Mask-RCNN~\cite{MaskRCNN}\\
        \bottomrule
    \end{tabular}
    }
    \label{tab:real_dataset}
\end{table}
\begin{figure}[t]
    \centering
    \includegraphics[width=0.95\linewidth]{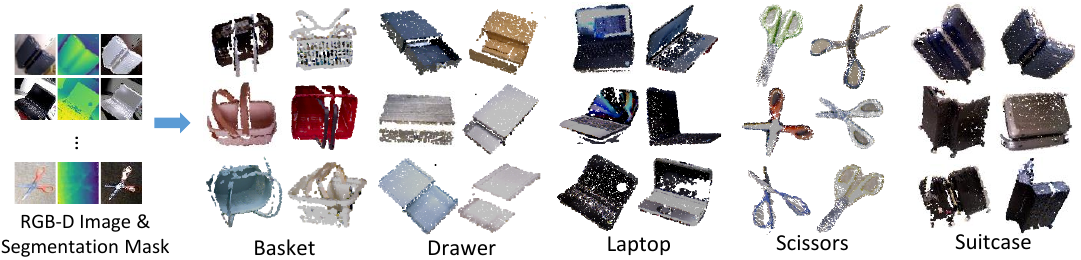}
    \caption{Example of object point cloud in the real-world dataset. We use RGB-D images and object segmentation masks to back-project object point cloud.}
    \label{fig:real_dataset}
\end{figure}

To evaluate the performance of \methodname\ in real-world scenarios, we introduce our novel real-world dataset. The real-world dataset contains 5 categories of articulated objects, \texttt{basket}, \texttt{laptop}, \texttt{suitcase}, \texttt{drawer}, and \texttt{scissors}, captured by ASUS Xtion RGB-D camera. For each category, we randomly select 4 objects for training and 2 objects for testing. For each object, we set 8 random joint states and captured about 30 frames of RGB-D images for each. We also generated object segmentation masks predicted with detection models such as Mask-RCNN~\cite{MaskRCNN} or Segment Anything Model~(SAM)~\cite{sam}. The object point cloud can be generated by combining the depth channel of RGB-D images with a segmentation mask. Tab.~\ref{tab:real_dataset} and Fig.~\ref{fig:real_dataset} show an overview of this dataset.
The annotation of the real-world dataset includes each part's segmentation, rotation, and translation and each joint's pivot and direction.

\section{Experiments}


\noindent \textbf{Datasets}: We use a synthetic dataset generated by authors of EAP~\cite{PLSEE} and our real-world dataset for evaluation. 
The synthetic dataset contains \texttt{laptop}, \texttt{safe}, \texttt{oven}, \texttt{washer}, and \texttt{eyeglasses} categories, selected from the mesh data in HOI4D~\cite{hoi4d} and Shape2Motion~\cite{shape2motion} dataset. We follow EAP~\cite{PLSEE}'s authors to render these mesh data into the partially observed point cloud, simulating the point cloud observation from a single-view camera.

\noindent \textbf{Baselines}: For the synthetic dataset, we choose EAP~\cite{PLSEE} and 3DGCN~\cite{3DGCN} as self-supervised and supervised method baselines. We also report the results of a ICP algorithm, and NPCS~\cite{ANCSH} with EPN~\cite{EPN} backbone, which the authors of EAP~\cite{PLSEE} implemented. For the real-world dataset, we trained 3DGCN~\cite{3DGCN} and PointNet++~\cite{pointnet++} as supervised method baselines.

\noindent \textbf{Evaluation Metrics}: For the synthetic dataset, we follow EAP~\cite{PLSEE} and report the mean values of segmentation IoU, part rotation error, part translation error, joint direction error, and the distance from a point to a line as joint pivot error.
For the real-world dataset, we follow category-level 6D object pose estimation methods~\cite{NOCS, GPV-pose, FS-net} and choose the mean average precision (mAP) with multiple thresholds. An instance's part pose is considered correct if the mean translation and rotation error of each part are both below the given thresholds. Specifically, we use thresholds $5,10,15$cm for translation, and $5^\circ,10^\circ,15^\circ$ for rotation. We also use the same thresholds for joint pivot and direction. For part segmentation, we use the mean value of intersection over union~(IoU) of each part and thresholds of $75\%, 50\%$ as metrics. 

\noindent \textbf{Evaluation Strategies}:
Because \methodname\ is a self-supervised model, it only predicts the relative poses of the input and the reconstruction instead of 
 the poses defined by humans.
Therefore, to evaluate our model's performance, we need to determine the poses of the reconstruction parts.
To achieve this, we follow EAP~\cite{PLSEE} and utilize ground truth labels from the training set.
In preparation, for each training data and each part, a relative pose between the reconstruction and the input is obtained through Equation~\ref{eqn:Z} by using a trained model, which, in combination with the ground truth pose, derives an estimated pose of each part of the reconstruction.
We use these estimated poses to determine one common pose for each part of the reconstruction via a RANSAC-based method. For evaluation, we use the common pose as the pose of each part of the reconstruction. See supplement material for more details.
We also note that for symmetric object categories such as \texttt{basket}, \texttt{laptop}, \texttt{scissors} and \texttt{suitcase}, the part segmentation is easily replaced with each other.
For each object, among all possible permutations of indices of segmentation labels, we choose the permutation with the largest mean IoU over parts.
The poses of parts are also permuted according to the chosen permutation.

\noindent \textbf{Training Settings}: 
We trained a model for each category for $20{,}000$ iterations with a batch size of $24$.
We used the Adam optimizer with a learning rate of $0.0001$ and halved the learning rate every $5{,}000$ iterations.
The total loss is defined as $\lambda_{\mathrm{o}} \mathcal{L}_{\mathrm{o}} + \lambda_{\mathrm{p}} \mathcal{L}_{\mathrm{p}} + \lambda_{\mathrm{regS}} \mathcal{L}_{\mathrm{regS}} + \lambda_{\mathrm{regD}} \mathcal{L}_{\mathrm{regD}} + \lambda_{\mathrm{regW}} \mathcal{L}_{\mathrm{regW}} + \lambda_{\mathrm{regP}} \mathcal{L}_{\mathrm{regP}} + \lambda_{\mathrm{regA}} \mathcal{L}_{\mathrm{regA}} + \lambda_{\mathrm{regJ}} \mathcal{L}_{\mathrm{regJ}}$, where $( \lambda_{\mathrm{o}}, \lambda_{\mathrm{p}}, \lambda_{\mathrm{regS}}, \lambda_{\mathrm{regD}}, \lambda_{\mathrm{regW}}, \lambda_{\mathrm{regP}}, \lambda_{\mathrm{regA}}, \lambda_{\mathrm{regJ}}) = (10, 10, 100, 10, 10, 10, 10, 10)$. We randomly sample 1024 points without RGB information from each object as input. 


\subsection{Results on the Synthetic Dataset}\label{sec: experiment_syn}

\begin{table}[tp]
    \caption{The mean metrics on partially observed point cloud from the synthetic dataset. \textit{Supervision} refers to the annotations used in training.} 
    \centering
    \scalebox{0.75}{
    \begin{tabular}{@{}ccccccccccc@{}}
        \toprule
        \multirow{2}{*}{Method} & \multicolumn{3}{c}{Supervision} & Segmentation & Rotation & Translation & Pivot & Direction & Memory & Speed\\
        & Pose & Segmentation & Joint & IoU↑ & (degree)↓ & ↓ & ↓ & (degree)↓ &(GB)↓ & (FPS)↑\\
        \cmidrule(lr){1-1} \cmidrule(lr){2-4} \cmidrule(lr){5-9} \cmidrule(lr){10-11}
        3DGCN~\cite{3DGCN}      &\checkmark &\checkmark &\checkmark  & \textbf{94.05} & 11.61 & \underline{0.093} & \textbf{0.084} & \underline{9.78} & - & -\\
        NPCS-EPN~\cite{ANCSH}   &\checkmark &\checkmark &\checkmark  & - & 11.05 & \textbf{0.080} & 0.147 & 15.20 & - & -\\
        ICP              && \checkmark &                     & 66.45 & 44.12 & 0.242 & - & - & - & -\\
        EAP~\cite{PLSEE}        &&&                     & 68.46 & \underline{10.44} & 0.121 & 0.162 & 23.09 & 9.23 & <1\\
        Ours                    &&&                     & \underline{80.70} & \textbf{8.10} & 0.129 & \underline{0.110} & \textbf{6.63} & \textbf{2.31} & \textbf{41}\\
        \bottomrule
    \end{tabular}
    }
    \label{tab:result_syn_partial_avg}
\end{table}

\begin{figure}[t]
    \centering
    \includegraphics[width=0.95\linewidth]{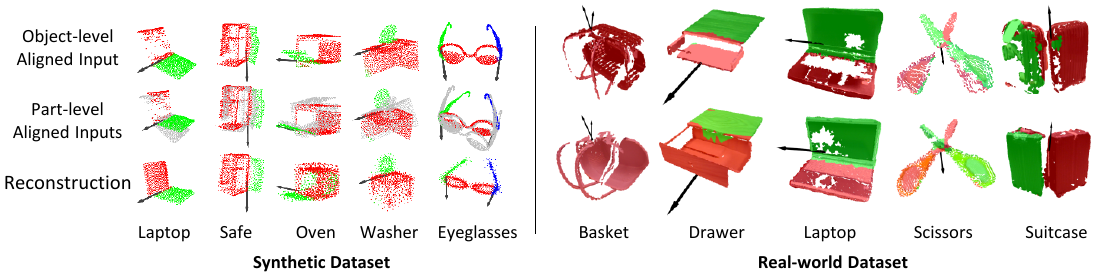}
    \caption{Visualization of object-level aligned inputs, part-level aligned inputs, and reconstructions of \methodname\ on the synthetic dataset (left) and two testing instances on the real-world dataset in each category (right). Segmentation is indicated by color, and joints are indicated by black arrow.}
    \label{fig:result_real}
\end{figure}

We compare the performance of \methodname\ on the partially observed point cloud from the synthetic dataset with other methods.
As the results in Tab.~\ref{tab:result_syn_partial_avg}, \methodname\ exceeds other self-supervised methods by a large margin on multiple metrics. These results show that \methodname\ can provide accurate joint and part pose prediction along with part segmentation.
The visualization shown in Fig.~\ref{fig:result_real} (left) demonstrates that object-level alignment can align the input with reconstruction holistically, and part-level alignment can align each part of the input with the corresponding part of the reconstruction.
Also, thanks to the object-level alignment for reducing the pose variance, our method achieved higher part segmentation performance when compared with EAP~\cite{PLSEE}.
However, the part segmentation performance still has room for improvement. Our assumption is that supervised 3DGCN~\cite{3DGCN} can directly learn segmentation with the geometric feature from the point cloud, while \methodname\ leverages the difference of point distance between each part-level aligned input and the reconstruction for the indirect learning of segmentation probability with $\mathcal{L}_{\mathrm{p}}$. Especially in the region close to the joint, where points of part-level aligned inputs easily overlap, the point distance had no significant difference between each part-level aligned input, resulting in sub-optimal segmentation performance.
We also compared our model in terms of inference speed and GPU memory with EAP~\cite{PLSEE}. \methodname\ utilizes less GPU memory and achieves faster inference speed.

\subsection{Results on the Real-world Dataset}\label{sec: experiment_real}

\begin{table}[tp]
    \caption{The comparison of mAP metrics on the real-world dataset. \textit{Supervision} refers to the annotations used in training.} 
    \centering
    \scalebox{0.75}{
    \begin{tabular}{@{}cccccccccccc@{}}
        \toprule
        \multirow{2}{*}{Method} & \multicolumn{3}{c}{Supervision} & \multicolumn{2}{c}{Segmentation↑} & \multicolumn{3}{c}{Joint↑} & \multicolumn{3}{c}{Part↑} \\
        & Pose & Segmentation & Joint & IoU75\% & IoU50\% & 5$^\circ$5cm & 10$^\circ$10cm & 15$^\circ$15cm & 5$^\circ$5cm & 10$^\circ$10cm & 15$^\circ$15cm\\
        \cmidrule(lr){1-1} \cmidrule(lr){2-4} \cmidrule(lr){5-6} \cmidrule(lr){7-9} \cmidrule(lr){10-12}
        3DGCN\cite{3DGCN}& \checkmark & \checkmark & \checkmark & \textbf{83.31} & \textbf{95.83} & \textbf{47.51} & \textbf{85.79} & \textbf{94.59} & \underline{13.07} & \textbf{46.77} & \textbf{68.66}\\
        PointNet++\cite{pointnet++}& \checkmark & \checkmark & \checkmark & 19.83 & 42.20 & \underline{21.06} & 57.38 & \underline{75.56} & 4.47 & 23.25 & 39.82\\
        Ours               & & &           & \underline{23.79} & \underline{50.42} & 12.57 & \underline{63.59} & 74.04 & \textbf{14.79} & \underline{46.09} & \underline{59.76}\\
        \bottomrule
    \end{tabular}
    }
    \label{tab:result_real_avg}
\end{table}

\begin{figure}[t]
    \centering
    \includegraphics[width=0.95\linewidth]{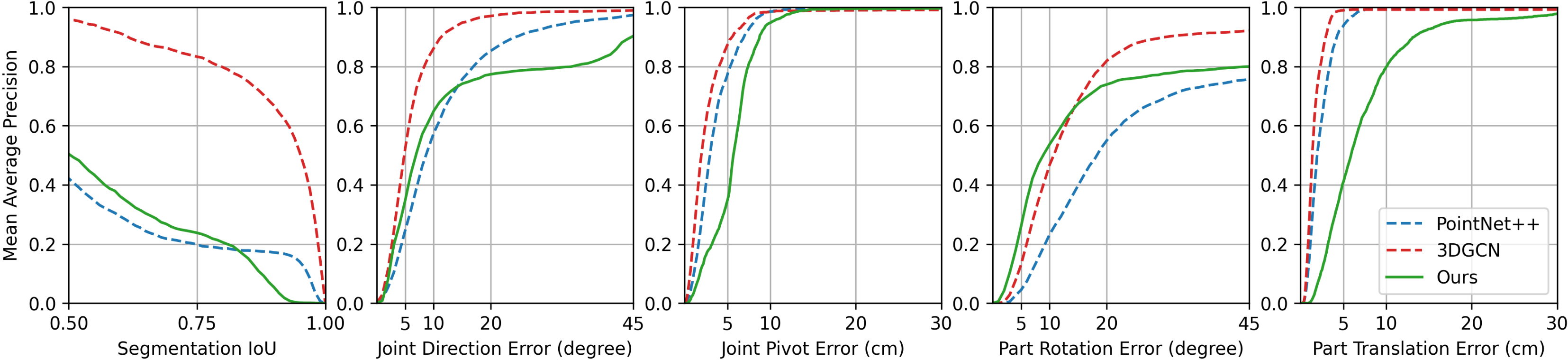}
    \caption{The comparison of mAP metrics on the real-world dataset.}
    \label{fig:result_line_mean}
\end{figure}

We conduct self-supervised training for \methodname\ and compared the result with supervised 3DGCN~\cite{3DGCN} and PointNet++~\cite{pointnet++} on the real-world dataset. The results are shown in Tab.~\ref{tab:result_real_avg} and Fig.~\ref{fig:result_line_mean} and the visualization is shown in Fig.~\ref{fig:result_real} (right).
\methodname\ achieves results better than or comparable to PointNet++~\cite{pointnet++} on all the metrics, and results comparable to 3DGCN~\cite{3DGCN} on part metrics, even without any annotations.
However, similar to the results on the synthetic dataset, part segmentation learning with $\mathcal{L}_{\mathrm{p}}$ requires accurate point distance between part-level aligned inputs and the reconstruction, which is extremely challenging in real-world environments where outliers and missing points commonly exist.
We also notice that our model still lags behind supervised methods in terms of the joint pivot and part translation metrics, as shown in \texttt{laptop} and \texttt{suitcase} visualization in Fig.~\ref{fig:result_real}. This phenomenon may be because the predicted joint pivots by our model, while capable of achieving part-level alignment, may not necessarily overlap with the actual joint pivots in reality. This also affects the performance of part translation based on joint movement.

\subsection{Ablation Studies}


\begin{figure}[t]
\begin{minipage}[b]{0.55\textwidth}
    \centering
    \scalebox{0.65}{
    \begin{tabular}{@{}ccccc@{\hskip 0.1in}ccc@{}}
        \toprule
        \multirow{2}{*}{\quad} & \multirow{2}{*}{$\mathcal{L}_{\mathrm{regS}}$} & \multirow{2}{*}{$\mathcal{L}_{\mathrm{regD}}$} & \multirow{2}{*}{$\mathcal{L}_{\mathrm{regW}}$} & \multirow{2}{*}{$\mathcal{L}_{\mathrm{regJ}}$} & Segmentation↑ & Joint↑ & Part↑ \\
         & & & &  & 50\% & 15$^\circ$15cm & 15$^\circ$15cm \\
        \cmidrule(lr){1-5} \cmidrule(lr){6-8}
        (a) &            & \checkmark & \checkmark & \checkmark & \textbf{51.87} & 42.79 & \underline{35.39}  \\
        (b) & \checkmark &            & \checkmark & \checkmark & 50.36 & \underline{51.52} & 32.15  \\
        (c) & \checkmark & \checkmark &            & \checkmark & 39.73 & 32.49 & 27.51 \\
        (d) & \checkmark & \checkmark & \checkmark &            & 47.52 & 36.27 & 20.49 \\
        Full& \checkmark & \checkmark & \checkmark & \checkmark & \underline{50.42}  & \textbf{74.04} & \textbf{59.76}  \\
        \bottomrule
    \end{tabular}
    }
    \captionof{table}{Results of ablation studies.} 
    \label{tab:result_ablation}
\end{minipage}
\hspace{3mm}%
\begin{minipage}[b]{0.4\textwidth}
    \centering%
    \includegraphics[width=\linewidth]{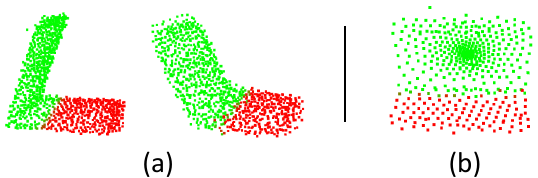}
    \caption{Reconstruction examples of ablation model (a) and (b).}
    \label{fig:ablation}
\end{minipage}
\end{figure}

We conduct four different ablation experiments on the real-world dataset, related to the shape variance regularization $\mathcal{L}_{\mathrm{regS}}$, the reconstruction density regularization $\mathcal{L}_{\mathrm{regD}}$, the segmentation regularization $\mathcal{L}_{\mathrm{regW}}$, and the joint pivot regularization $\mathcal{L}_{\mathrm{regJ}}$, as shown in Tab.~\ref{tab:result_ablation}. Examples of the reconstructions of objects in the real-world dataset \texttt{laptop} category are shown in Fig.~\ref{fig:ablation}. As the reconstructions and performance of ablation model (a) show, without $\mathcal{L}_{\mathrm{regS}}$, the reconstructions' joint state is not fixed, which results in a huge performance drop at metrics of joint and part prediction. For ablation model (b), without $\mathcal{L}_{\mathrm{regD}}$, reconstruction's points are concentrated into a small region, which affects the overall performance of our model. For ablation model (c), without $\mathcal{L}_{\mathrm{regW}}$, some objects are regarded as single-part objects, and we fail to generate valid joint parameters. Finally for ablation model (d), without $\mathcal{L}_{\mathrm{regJ}}$, joint pivot may be placed outside of the object, resulting in poor performance on both joint and part pose metrics.

\section{Failure Cases and Limitations}
\textbf{Failure Cases:}
We found that \methodname\ fails for objects belonging to categories where some parts comprise only a small fraction of the entire object and their movement does not significantly affect the overall shape. For \texttt{basket} category, as shown in Fig.~\ref{fig:result_real}, the handle parts account for 16.9\% of the entire object (median in the testing set) and the movement of these parts results in small changes to the overall shape. This means that even without part-level alignment, our canonical reconstruction is sufficiently close to the overall object. This also leads to our model's inability to correctly segment parts and predict joint movements.

\noindent \textbf{Limitations:}
\methodname\ requires the number of parts and joint types as known information, which limits its ability to learn from objects in categories with unknown joint types or variable numbers of joints and parts.

\section{Conclusion}

We proposed a novel approach, \methodname, and a new real-world dataset for the self-supervised category-level articulated object pose estimation. Our approach achieves state-of-the-art performance among self-supervised methods and comparable performance to previous supervised methods, yet with real-time inference speed. Our future plan is to design a self-supervised universal pose estimation model, which can be trained with inner-category data and automatically detect the number of parts, number of joints, and joint type. 

\textbf{Acknowledgements:} We thank Ryutaro Yamauchi and Tatsushi Matsubayashi from ALBERT Inc. (now Accenture Japan Ltd.) for their insightful suggestions and support. This work was supported by JST FOREST Program, Grant Number JPMJFR206H.


%
%
\bibliographystyle{splncs04}
\bibliography{main}
\end{document}